\documentclass{article}

\pdfoutput=1
\usepackage{arxiv}

\usepackage[utf8]{inputenc} 
\usepackage[T1]{fontenc}    
\usepackage{hyperref}       
\usepackage{url}            
\usepackage{booktabs}       
\usepackage{amsfonts}       
\usepackage{nicefrac}       
\usepackage{graphicx}
\usepackage{doi}
\usepackage{lmodern}
\usepackage{mathtools}
\usepackage{multirow}

\title{Semi-Supervised Cascaded Clustering for Classification of Noisy Label Data}

\author{
 Ashit Gupta \\
  TCS Research\\
  Tata Consultancy Services Ltd.\\
  Pune, Maharashtra, India \\
  \texttt{ashit.gupta@tcs.com} \\
   \And
 Anirudh Deodhar \\
  TCS Research\\
  Tata Consultancy Services Ltd.\\
  Pune, Maharashtra, India \\
  \texttt{anirudh.deodhar@tcs.com} \\
  \And
 Tathagata Mukherjee \\
  TCS Research\\
  Tata Consultancy Services Ltd.\\
  Pune, Maharashtra, India \\
  \texttt{m.tathagata@tcs.com} \\
  \And
 Venkataramana Runkana\\
  TCS Research\\
  Tata Consultancy Services Ltd.\\
  Pune, Maharashtra, India \\
  \texttt{venkat.runkana@tcs.com} \\
}

\begin{document}
\maketitle
\begin{abstract}
The performance of supervised classification techniques often deteriorates when the data has noisy labels. Even the semi-
supervised classification approaches have largely focused only on the problem of handling missing labels. Most of the approaches addressing the noisy-label data rely on deep neural networks (DNN) that require huge datasets for classification tasks. This poses a serious challenge especially in process and manufacturing industries, where the data is limited and labels are noisy. We propose a semi-supervised cascaded clustering
(SSCC) algorithm to extract patterns and generate a cascaded tree of classes in such datasets. A novel cluster evaluation matrix (CEM) with configurable hyperparameters is introduced to localize and eliminate the noisy labels and invoke a pruning criterion on cascaded clustering. The algorithm reduces the dependency on expensive human expertise for assessing the accuracy of labels. A classifier generated based on SSCC is found to be accurate and consistent even when trained on noisy label datasets. It performed better in comparison with the support vector machines (SVM) when tested on multiple noisy-label datasets, including an industrial dataset. The proposed approach can be effectively used for deriving actionable insights in industrial settings with minimal human expertise.
\end{abstract}

\keywords{Semi-supervised cascaded clustering \and Classification \and Noisy-label datasets \and Industrial use case}


\section{Introduction}
Data mining and pattern recognition have enabled businesses to be proactive and take smart decisions based on data-derived knowledge over past two decades. The most common and perhaps the most challenging type of application is classification that can range from binary to multiclass classification \cite{JAIN01}. Supervised classification techniques work well on data rich in quality and quantity, but the performance declines with drop in richness of data. The issue could be due to lack of sufficient data or mislabeling of data. Unsupervised clustering may help solve some of the issues and assist in identifying patterns from the datasets. Although they are effective in checking for coherence in the data, clustering models are not indicative in nature and lack specific outcome measures. Semi-supervised classification has emerged as a viable alternative for learning using partially labeled data \cite{SCHWENKER01}. These algorithms exploit the presence of unlabeled data to enhance the accuracy of initial classifier built using labeled data. Some of the approaches include self-training \cite{SCUDDER01}, co-training \cite{BLUM01}, transductive support vector machines \cite{Joachin01}, generative models \cite{NIGAM01}, and graph based methods \cite{BELKIM01}. Self-training initially trains a supervised classifier using only the labeled data and then identifies a label for unlabeled data using the classifier. The high confidence labeled data is used again to re-train the classifier. This iterative process yields a better classifier model than the initial one. However, the accuracy of final classifier often deteriorates when the data space occupied by the labeled data is not representative of the entire data space. Semi-supervised clustering has also been proposed where the clustering algorithm was provided with observation linking constraints to force specific observations to fall under one cluster or separate out into different clusters \cite{BASU01,SHENTAL01}. However, predefining the constraints may not always be possible. Utilizing clustering outcomes to improve the performance of semi supervised classification has also been proposed previously \cite{SCHWENKER01,Pironn01,GAN01,Peikari01}.

Although handling of missing labels in the data during classification and improving the classification accuracy using clustering has been effectively addressed by methods mentioned above, there is relatively limited work on handling noisy-label data \cite{WU01,Li01,WEI01}. Noisy-label data is hard to spot and when used for guiding the classification/clustering in the semi-supervised classification techniques, may hamper the learning of the underlying data space. Secondly, despite recent advances in Automated Machine Learning (AutoML) \cite{Thorton01,Feurer01}, especially for supervised learning, studies on automation of semi-supervised classification are limited \cite{YU01}. 

Another branch of methods trains deep neural networks (DNN) for learning noisy labels from the data. Zhang et al. showed that the DNN often overfits the noisy labels and subsequently the performance drops \cite{Zhang01}. Therefore, efforts are required to remove the noisy labels from the dataset. Most of the methods for learning noisy labels take the loss correction approach. A few methods estimate the noise transition matrix and use it to correct loss functions \cite{Patrini01,Goldberger01}. Estimation of such transition matrix is often challenging. Other methods correct the loss function either explicitly or implicitly and relabel the noisy data \cite{VEIT01,LEE01}. These methods require access to a set of correctly labeled data, which might not always be possible. State-of-the-art methods like dividemix \& SemiNLL discard the labels likely to be noisy and leverage the noisy samples as unlabeled data to regularize the model from overfitting and improve generalization performance \cite{Dividemix01,Seminll01}. However, aforementioned DNN based methods typically require large amount of data to train from, which may present a challenge particularly in process and manufacturing industrial settings, where the data lacks in both quality and quantity. 

We address the issue of classifying a set of partially mislabeled or noisy label data using a semi-supervised approach. We propose a distance-based automated semi-supervised cascaded clustering algorithm (SSCC) to identify the correct patterns/classes in the data space while eliminating the possible noisy-label information. The algorithm is supported by a novel Cluster Evaluation Matrix (CEM) with configurable hyperparameters. The matrix serves as a guide for evaluating the identified clusters, separating noisy-label data and pruning the cascaded clusters.  Further, a framework to generate a self-trained classifier is presented, that inherits the architecture of SSCC and trains well even on noisy-label data, when compared with modern supervised classifiers such as support vector machine (SVM). The proposed method is developed as a part of a digital twin solution for a thermal power plant, for real-time detection and classification of coal type. The coal classification assists in real-time optimum operation of the plant and reduction in greenhouse emissions. The proposed method performs well even on limited noisy data and is tested with multiple such datasets. 

\pagestyle{fancy}
\fancyhead[L]{Semi-Supervised Cascaded Clustering for Classification of Noisy Label Data}

\section{Semi-Supervised Cascaded Clustering}
\label{sec:headings}
The clustering process for a set of data with 'n' features can be considered as observing the data from a frame of reference with n-coordinates, aggregating the points close to each other and separating the points that are away from each other. All separation and aggregation may not be possible by looking at the data from only one frame of reference in a single go. The core idea behind a cascaded clustering algorithm is viewing and separating the data at multiple levels while maximizing the clustering effectiveness at every level by changing the frame of reference and number of clusters. The algorithm is automated by measuring the clustering effectiveness at every level using the available knowledge through labels and identifying the stopping criterion via a clustering evaluation matrix. Algorithm-1 describes the formation of cascaded structure using the SSCC algorithm.

\begin{table}[h]
	\label{algo:clustering}
	\begin{tabular*}{\textwidth}{l l l l l l}
		\hline
		\multicolumn{4}{c}{\textbf{Algorithm-1}} : SSCC& & \\
		\hline
		& \multicolumn{4}{l}{\textbf{input} ~~ :  labeled data (LD),}  &\\
		& & \multicolumn{3}{l}{ ~ Hyperparameters ($\lambda_{CEM}$, $\lambda_{CS}$, $\lambda_{OL}$).} & \\
		& \multicolumn{4}{l}{\textbf{output} ~: Cascaded clustering table, final classes.} & \\
		& \textbf{begin} & & && \\
		1 & \multicolumn{3}{l}{~~\vline  ~~ Normalize the data.} &&\\
		2 & \multicolumn{3}{l}{~~\vline  ~~ Set level v=0.} &&\\
		3 & \multicolumn{5}{l}{~~\vline  ~~ Generate unique sets of features ($S$), $S=\{[x_1,x_3],[x_1,x_2,x_4],...,\}$. } \\
		4 & \multicolumn{5}{l}{~~\vline  ~~ Cluster all sets based on best Silhouette score, ($S$).} \\
		5 & \multicolumn{5}{l}{~~\vline  ~~ Calculate Completeness score for all sets, ($S$).} \\
		6 & \multicolumn{5}{l}{~~\vline  ~~ Select the best clustering result among ($S$)  with max completeness score ($CS_{max}$).} \\
		7 & \multicolumn{5}{l}{~~\vline  ~~  Set K=Number of cluster in the best clustering result.} \\
		8 & \multicolumn{5}{l}{~~\vline  ~~  Set C=Number of labels in the best clustering result.} \\
		9 & \multicolumn{5}{l}{~~\vline  ~~  If ($CS_{max}$< $\lambda_{CS}$)} \\
		& \multicolumn{5}{l}{~~\vline  ~~ ~~  Assign the data at level v as a class. } \\
		& \multicolumn{5}{l}{~~\vline  ~~ ~~ Label the class obtained.} \\
		& \multicolumn{5}{l}{~~\vline  ~~ Else} \\
		& \multicolumn{5}{l}{~~\vline  ~~ ~~ Compute CEM for best clustering result.} \\
		& \multicolumn{5}{l}{~~\vline  ~~ ~~ For j in K do} \\
		& \multicolumn{5}{l}{~~\vline  ~~ ~~  ~~ ~~\vline ~~ For i in C do} \\
		& \multicolumn{5}{l}{~~\vline  ~~ ~~ ~~ ~~\vline ~~ ~~ ~~\vline ~~ If (CEM[i,j] $ < \lambda_{OL}$) do} \\
		& \multicolumn{5}{l}{~~\vline  ~~ ~~ ~~ ~~\vline  ~~ ~~ ~~\vline ~~ ~~ ~~ Remove associated data as possible noisy data.} \\
		& \multicolumn{5}{l}{~~\vline  ~~ ~~ ~~ ~~\vline  ~~ ~~ ~~\vline ~~  If (CEM[i,j] $ \ge \lambda_{CEM}$) do} \\
		& \multicolumn{5}{l}{~~\vline  ~~ ~~ ~~ ~~\vline ~~ ~~ ~~\vline ~~ ~~ ~~ Term Cluster (j) as a class.} \\
		& \multicolumn{5}{l}{~~\vline  ~~ ~~ ~~ ~~\vline  ~~ ~~ ~~\vline ~~ ~~ ~~  Label the class obtained.} \\
		& \multicolumn{5}{l}{~~\vline  ~~ ~~ ~~ ~~\vline   ~~ ~~ ~~\vline ~~ Else do} \\
		& \multicolumn{5}{l}{~~\vline  ~~ ~~ ~~ ~~\vline  ~~ ~~ ~~\vline ~~ ~~ ~~  do v=v+1 } \\
		& \multicolumn{5}{l}{~~\vline  ~~ ~~ ~~ ~~\vline  ~~ ~~ ~~\vline ~~ ~~ ~~  Repeat from Step 3 for cascading the cluster j.} \\
		& \multicolumn{5}{l}{~~\vline  ~~ ~~  ~~ ~~\vline ~~ ~~ end} \\
		& \multicolumn{5}{l}{~~\vline  ~~ ~~ ~~ end} \\
		& \textbf{end} &&&& \\
		\hline
	\end{tabular*}
\end{table}

The first step is to normalize the data using mean and standard deviation. This is only done once i.e. at cascade level 0. Unsupervised distance-based clustering algorithm such as k-means or k-medoids is used on the labeled observations \cite{Alsabti01,SHENG01}. Although the observations have labels, they are not considered at this point. The algorithm generates multiple sets of combinations of features or multi-dimensional frames of reference (S=$\{[x_{1},x_{2},x_{3}],[x_{1},x_{2},x_{4},x_{6}],....\}$). We use Silhouette score \cite{Zhou01} for identifying the optimum number of clusters for a given combination of features (from the set S), with an assumption that the maximum number of clusters cannot exceed the number of labels present at the cascade level \cite{Zhou01}. Semi-supervised procedure of training the model is initiated by computing the completeness score for each of the clustering outcomes with all list elements present in sets S using the available labels \cite{Rosenberg01}. A higher completeness score indicates that majority number of observations under the same label are assigned to a single cluster. This step ensures meaningful separation of data and removal of possible outliers and data associated with incorrect labels. Another important measure of semi-supervised clustering, homogeneity \cite{Rosenberg01} is not invoked at this stage because the cascading algorithm assumes there is a scope for two or more labels forming a single class (or a cluster). The mathematical formulation of the Completeness score is given below.
\begin{equation}
H(K|C) = - \sum_{c=1}^{\left|C\right|} \sum_{k=1}^{\left|K\right|} \frac{n_{c,k}}{N} \log \frac{n_{c,k}}{\sum_{k=1}^{\left|K\right|} n_{c,k}} 
\end{equation}
\begin{equation}
H(K,C) = - \sum_{k=1}^{\left|K\right|} \sum_{c=1}^{\left|C\right|} \frac{n_{c,k}}{N} \log \frac{n_{c,k}}{N} 
\end{equation}		
\begin{equation}
{c}  =  \begin{cases} 
1   & if ~ H(K,C) = 0 \\
1 - \frac{H(K|C)}{H(K,C)} & else 
\end{cases}
\end{equation}

Where, K is total number of clusters, C is total number of labels, H(K/C) represents conditional entropy, $n_{c,k}$ is the value of an element for a given cluster number and the label number, N are the number of data points present at a cascade level and c is the completeness score. This nomenclature is used across the paper. The clustering result (among S combinations) that has the highest completeness score $(CS_{max})$ is selected as the best result and $CS_{max}$ is compared against a threshold ($\lambda_{CS}$). This threshold is treated as a configurable hyperparameter. If the maximum completeness score is less than the threshold ($\lambda_{CS}$), it indicates that the clustering at that level cannot separate the observations meaningfully and hence no further clustering needs to be done. Entire data used for clustering at this stage is then assigned to a single class. On the other hand, if the maximum completeness score is higher than the threshold ($\lambda_{CS}$), it indicates that clustering was able to split the observations meaningfully. However, the decision on whether further clustering is required or not cannot be taken based on the completeness score alone. Here, we propose a novel matrix called Cluster Evaluation Matrix (CEM) that helps in deciding whether a particular cluster should be separated further or not.

The Cluster Evaluation Matrix (CEM) enables determination of a cluster as a class and separation of outliers of a class (probable noisy-label observations). It serves as a stopping criterion for pruning of the overall cascading architecture. The elements of the CEM are computed using the equation below. 

\begin{multline}
A=\frac{n_{c,k}}{\sum_{c=1}^{\left|C\right|} n_{c,k}} \max \bigg(\frac{\log \left( n_{c,k} + 1 \right)}{\log \left( \sum_{k=1}^{\left|K\right|} n_{c,k} + 1 \right)},
\exp^{\left( \frac{n_{c,k}}{\sum_{k=1}^{\left|K\right|} n_{c,k}} - 1 \right)}  \bigg) 
\end{multline} 

The dimensions of the generated CEM are [K,C]. Often the observations belonging to a specific label get separated into multiple clusters during clustering, due to outliers and noisy-label data under that label. The relative proportion of how the observations are separated across different clusters and the number of observations separated in a given cluster are two important aspects for deciding on continuing the clustering process. The first logarithmic term inside the max function in equation 4 accounts for the total number of observations associated with a label in each cluster. The second exponential term in the max function accounts for the relative proportion of observations associated with that label across different clusters. Taking the maximum of the aforementioned terms avoid false alarms of outliers or noisy labels in the cluster evaluation matrix.The term outside the max function provides a measure of the dominance of a particular label in the given cluster.

\begin{figure}[h]
	\centering
	\includegraphics[scale=0.7]{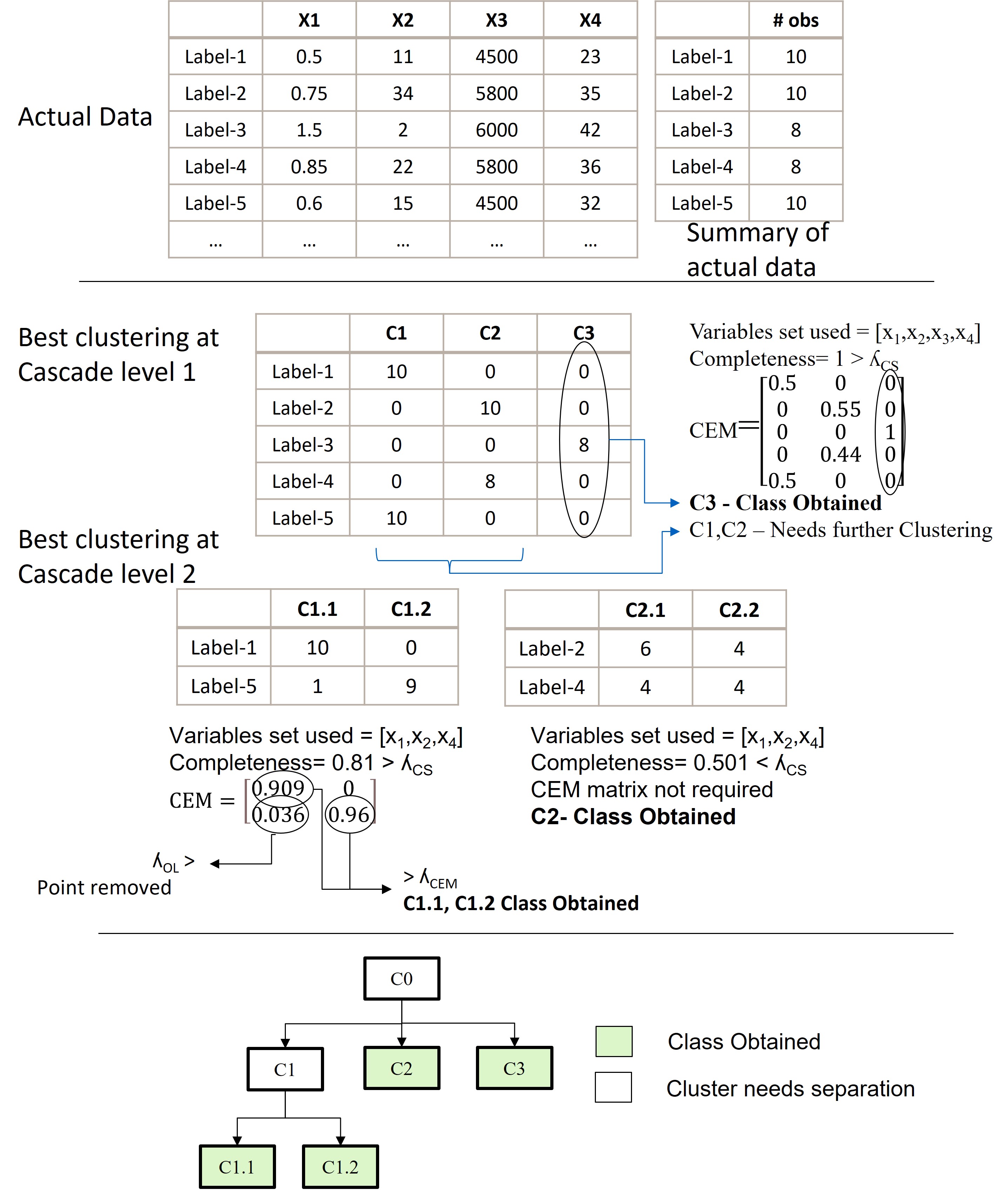}
	\caption{Illustrative example of cascaded clustering.}
	\label{figure:illustration}
\end{figure}

The two configurable hyperparameter thresholds $\lambda_{CEM}$ and $\lambda_{OL}$ are designed for decision making, which can be fine tuned based on the nature of dataset, prior knowledge about labels and the desired number of classes. If any element of the CEM is less than $\lambda_{OL}$, then the observations corresponding to that cluster and label are termed as outliers and are removed from subsequent clustering. Usually, these are the noisy-label observations associated with each of the labels. Then for each cluster, all the elements of CEM are checked against $\lambda_{CEM}$. If any element of a chosen cluster is greater than $\lambda_{CEM}$, that cluster is designated as a class. The calculation of CEM is done in such a way that it remains true irrespective of values of other elements in that cluster. If no value in a given cluster of CEM is greater than $\lambda_{CEM}$, the cluster is assigned for further clustering. The observations associated with that cluster are then taken forward in a cascaded manner for re-clustering using the same methodology. The cascading of clusters and their re-clustering creates a tree like structure until all branches of the tree are halted based on the completeness score and CEM. The bottom-most or leaf nodes of the tree are the classes obtained from the data. An illustration of SSCC (Algorithm 1) is shown with an example in figure ~\ref{figure:illustration}.

\section{Self-generated cascaded Classifier}
Once the final tree from the cascaded clustering is obtained, a self-trained classifier is derived from the tree architecture. The classifying method follows the exact path followed by an observation in the training set, as shown in algorithm 2.  The training set refers to a set of observations used for cascaded clustering. First, the new observation to be classified is normalized using the normalization scale of the training set. Then, at each level of classification, the nearest cluster from the set of cascaded clusters to the observation is identified. This is done using any distance-based metric such as the Euclidean distance, specifically calculated based on the features used at that level of clustering only. A central measure such as mean or median of each cluster is used to calculate this distance. The choice of distance and central measure can vary depending on the distribution of data and the clustering technique used during cascaded clustering. This process is repeated at each level until the nearest class is identified for the new observation. If the new observation is found to be dissimilar with respect to the existing classes or clusters at any level, a new label is assigned to it temporarily. If sufficient number of such new labels are obtained at any point during the classification exercise, a new class can be formulated by using the SSCC by combining the initial training set and the new labeled test set used during classification. Algorithm-2 describes the complete functioning of the SSCC derived classifier. The classifier inherits the benefits of SSCC and hence is likely to perform better over noisy label datasets compared to conventional machine learning classifiers. This is discussed in the next section. 

\begin{table}[h]
	\begin{tabular*}{\textwidth}{l l l l l l}
		\hline
		\multicolumn{4}{c}{\textbf{Algorithm-2}} : SSCC Classification & & \\
		\hline
		& \multicolumn{4}{l}{\textbf{input} ~~ :   $X=\{x_i\}$ }  &\\
		& \multicolumn{4}{l}{\textbf{model} ~~:    SSCC model }  &\\
		& \multicolumn{4}{l}{\textbf{output} ~: Representative Class (CL) } & \\
		& \textbf{begin} & & && \\
		1 & \multicolumn{5}{l}{~~\vline  ~~ Normalize vector X with normalization used for SSCC.} \\
		2 & \multicolumn{5}{l}{~~\vline   ~~ Retrieve cascade level L from SSCC model } \\
		3 & \multicolumn{5}{l}{~~\vline   ~~ For $v$ in L(Cascade levels) do } \\
		& \multicolumn{5}{l}{~~\vline   ~~  ~~ ~~\vline ~~ Choose the same features as were for SSCC for $v^{th}$ level; $X{[x_i]}$.} \\
		& \multicolumn{5}{l}{~~\vline   ~~  ~~ ~~\vline ~~ Retrieve K= number of clusters at level $v$.} \\
		& \multicolumn{5}{l}{~~\vline   ~~  ~~ ~~\vline ~~ Retrieve cluster center $A_{j,v}$; $j$ $\in$ [1,K]} \\
		& \multicolumn{5}{l}{~~\vline   ~~  ~~ ~~\vline ~~ Calculate Euclidean distance, $d_{j,v}=$ ED$(X, A_{j,v})$.} \\
		& \multicolumn{5}{l}{~~\vline   ~~  ~~ ~~\vline ~~ Find minimum $d_{j,v}$, assign name “$j$” to cluster with $min$ $d_{j,v}$.} \\
		& \multicolumn{5}{l}{~~\vline   ~~ ~~ ~~\vline ~~ If j is Class (CL) do } \\
		& \multicolumn{5}{l}{~~\vline   ~~ ~~ ~~\vline ~~ ~~ ~~ Classify X to j class.  } \\
		& \multicolumn{5}{l}{~~\vline   ~~ ~~ ~~\vline ~~ ~~ ~~ Break   } \\
		& \multicolumn{5}{l}{~~\vline   ~~ ~~ ~~\vline  ~~ Else do  } \\
		& \multicolumn{5}{l}{~~\vline   ~~ ~~ ~~\vline ~~ ~~ ~~ Subset the cascaded clusters under cluster "j" with iterating over $v$.} \\
		& \multicolumn{5}{l}{~~\vline   ~~ ~~ end} \\
		& \textbf{end} & & && \\
		\hline
	\end{tabular*}
	\label{algo:classification}
\end{table}  

\section{Results \& Discussions.}
We determined the effectiveness of SSCC algorithm and the generated cascaded classification model on four datasets available in the public domain as shown in table ~\ref{table:Experimental}. Each dataset was divided into train (for SSCC 85\%-90\%) and test (for cascaded classifier 10\%-15\%) data. A 10\% to 30\% noisy-labels were introduced randomly into the training data for all the datasets. The test data for each dataset was free from noisy-labels and was kept constant across the variations of datasets. The SSCC clustering was done on the training data for each dataset and the corresponding classifier was run on the respective test dataset. A very well-known classifier, Support Vector Machine (SVM) was used for benchmarking the accuracy, given its capability to adapt to different sized datasets \cite{Lee02}. A comparison of the SVM classifier against the SSCC classifier on the selected small datasets is presented later.

\begin{table}[h]
	\small
	\begin{center}
		\caption{Experimental dataset description.}
		\begin{tabular}{l c c c}
			\hline
			Datasets & \#Observations & \#Labels & \#Features\\
			\hline
			Coal \cite{Huleatt01} & 370 & 37 & 6  \\
			Ecoli \cite{Dua01,Nakai01,Nakai02} & 337 & 7 & 7 \\
			Wine \cite{Dua01} & 179 & 3 &  13  \\
			Eucalyptus \cite{Thomson01,Bulloch01} & 680 & 13 & 12 \\
			\hline
		\end{tabular}
		\label{table:Experimental}
	\end{center}
\end{table}  

\subsection{Semi-supervised Cascaded Clustering (SSCC)}
The SSCC algorithm was run on all the perfectly labeled (0\% noisy-label data) and noisy-labeled variations (10\%-30\% noisy-label data) of all selected datasets. Two distance based clustering algorithms i.e. k-means and k-medoids were used for all the variations of data. On the basis of experiments conducted, the default values of threshold parameters were found be as follows: $\lambda_{CS}$ =.85 to 0.9, $\lambda_{CEM}= 0.9$ and $\lambda_{CS} =0.01$ for the coal dataset. Table ~\ref{table:threshold} in Appendix A presents the values of threshold $\lambda$ for all datasets. Higher values of $\lambda_{CS}$ and $\lambda_{CEM}$ represent stringent condition on satisfying clustering criteria, whereas a higher value of $\lambda_{OL}$ indicates more data can be removed as an outlier. As a simplified illustration of the logic, consider a simple example of 3 cities i.e. Los Angeles, Washington DC and New York and their respective locations. Someone might want to classify Washington DC \& New York under one cluster and Los Angeles in another (based on their location) or some experiment might need all three to be in separate clusters. In the former case the values of threshold parameters determining the class ($\lambda_{CS}$ and $\lambda_{CEM}$) would be higher than the later. In the latter case, the segregation would require two cascaded levels to segregate Washington DC and New York in the second level. This configuration also helps the user to restrict the number of clusters depending upon the granularity required, which is an essential requirement in industrial settings for ease of interpretation. 

\begin{figure}[h]
	\centering
	\includegraphics[scale=1.0]{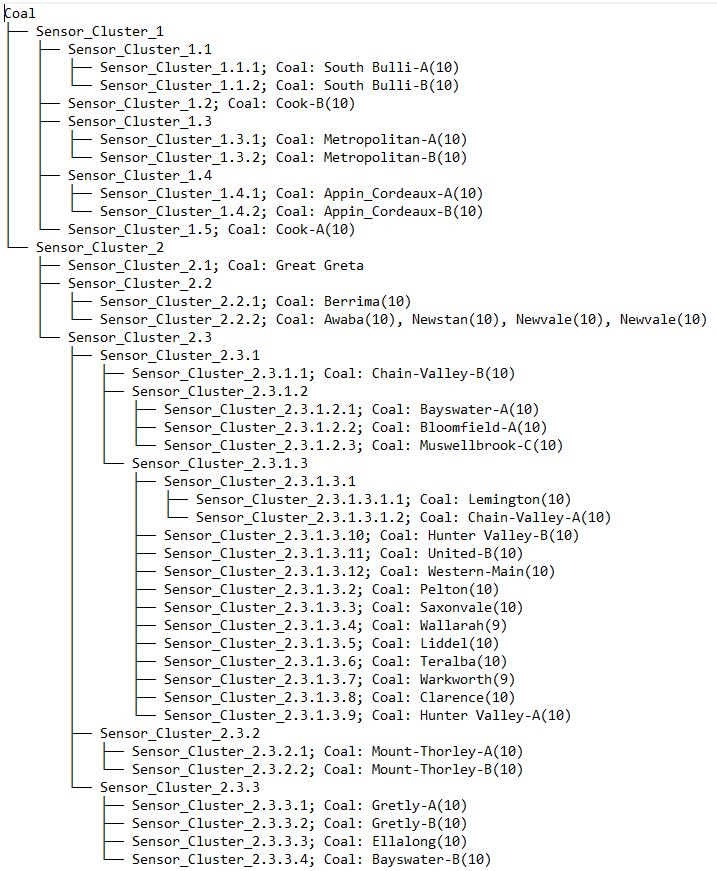}
	\caption{Outcome of SSCC on perfectly labeled coal dataset.}
	\label{figure:coaltable}
\end{figure}

A representative outcome of SSCC algorithm when used on perfectly labeled coal dataset is shown in figure ~\ref{figure:coaltable}. The cascaded architecture of the final clustering tree is visible with all the labels (coal names) associated with each class (leaf nodes of the tree). The algorithm automatically stops at the shown leaf nodes of the tree based on the completeness and CEM criteria. The detailed results from the SSCC algorithm are provided with the supplementary material. This tree was then used to generate a cascaded classifier and was run on the test data. Similar types of cascading architectures were obtained for each of the perfectly labeled and noisy-label datasets of all types. 

\subsection{SSCC classification}
Table~\ref{table:results} presents the comparison of classification results using cascaded classifier and SVM on all variations of datasets, in terms of percentage accuracy. The quality of SVM classifier deteriorated with addition of noisy-labels in the training data, in almost all experiments. In case of coal dataset, both train and validation accuracy dropped as the percentage of noisy label information increased. Similar drop in accuracy was observed for all other test datasets for noisy-label data compared with the corresponding perfectly labeled data, with an exception of wine dataset. In wine dataset both the models did not show any drop in accuracy due to limited number of test observation and relatively shorter size of the dataset. However, in all the other datasets, the SSCC classification algorithm tends to provide accurate and consistent results with noisy-label information as well. This is because the SSCC algorithm filtered out the noisy information through the CS and CEM criteria and the cascaded architecture. 

The algorithm proposed can solve general classification problems in industrial datasets where the labeled data lacks both in quantity and quality. The novelty of our algorithm lies in the usage of cascaded clustering and a set of thresholds that assist in removing outliers and incorrectly labeled information automatically. The cascading structure also generates a classifier which is free from labeling biases and is suitable for online learning using real-time data. This concept has been successfully applied for classification of small incorrectly labeled datasets, typically found in industrial settings.

The SSCC algorithm is relevant for many industrial use cases, for example, classification of coal types in thermal power plants, where coal is burnt to produce power. Correct classification of the type of incoming coal in terms of its properties and composition can greatly assist in maximizing plant efficiency, reducing emissions of pollutants and in improving the health of the plant. The number of coal labels is large and often, there are noisy-labels in the recorded coal types. The SSCC algorithm when tested on coal dataset, was able to classify coals into appropriate classes. The number of classes generated can be controlled by tuning the SSCC hyperparameters. The algorithm also successfully removed the noisy-label coal observations during clustering and hence performed perfectly when tested as a classifier on new test dataset. The SSCC algorithm can be utilized for online clustering and real-time classification of coals or any other materials in industrial settings. Since the algorithm is generic, it can be effectively used for any scenario in real world where classification is important, and the data is likely to have noisy labels.

\begin{table}[h]
	\small
	\centering
	\caption{ Comparison of Cascaded Classifier results with SVM on multiple datasets.}
	\begin{tabular}{l c c c c c c c}
		\hline
		Datasets & Mislabeling & SSCC kmeans & SSCC kmedoids & \multicolumn{3}{c}{SVM} \\
		& (\%)& (\% accuracy) & (\% accuracy) & \multicolumn{3}{c}{(\% accuracy)}\\
		\cline{5-7}
		& & & & Train & Validation & Test\\
		\hline
		\multirow{4}{*}{Coal Data} & 0 & 100 & 100 & 100 & 100 & 100 \\
		& 10 & 100 & 100 & 81.7 & 86.8 & 92.6 \\
		& 20 & 100 & 100 & 76.4 & 82.4 & 89 \\
		& 30 & 100 & 100 & 68.9 & 67.6 & 86.4\\
		\hline
		\multirow{4}{*}{Eucalyptus Data} & 0 & 99.1 & 99.1 & 89.6 & 81.5 & 90.1 \\
		& 10 & 97.5 & 99.1 & 80.7 & 77.2 & 83.7 \\
		& 20 & 98.34 & 99.1 & 69.7 & 75.5 & 77.6 \\
		& 30 & 99.1 & 99.1 & 59.6 & 69.8 & 75.2\\
		\hline
		\multirow{4}{*}{Wine Data} & 0 & 100 & 100 & 99.7 & 100 & 100 \\
		& 10 & 100 & 100 & 86.5 & 81.2 & 100 \\
		& 20 & 100 & 100 & 77.3 & 81.2 & 100 \\
		& 30 & 100 & 100 & 67.4 & 75 & 100\\
		\hline
		\multirow{4}{*}{Ecoli Data} & 0 & 77.3 & 81.1 & 90.7 & 88.2 & 77.3 \\
		& 10 & 77.3 & 81.1 & 84.8 & 93.8 & 72.7 \\
		& 20 & 77.3 & 81.1 & 75.9 & 81.2 & 72.7 \\
		& 30 & 77.3 & 81.1 & 65.6 & 71.9 & 68.2\\

		\hline
	\end{tabular}
	\label{table:results}
	
\end{table}  

\subsection{Possible Improvements.}
Although, we have demonstrated SSCC algorithm using k-means and k-medoids clustering algorithms and using Euclidean distance metric for classification, it can be easily substituted with equivalent metric depending upon the application and data. Further exploration is needed to address noisy-label datasets where distance-based clustering may not be applicable. The SSCC algorithm can be improved further by optimizing the feature selection at every cascaded level of clustering, as the current method is computationally expensive, especially for large number of features. In addition, the classifier algorithm can be improved by provisioning for addition of new classes automatically as well as robust outlier detection at every cascaded level. Selection of hyperparameters for SSCC clustering varies with respect to data, domain, and desirable outcomes. Further work is required to automate it.

\section{Conclusion}
An automated semi-supervised cascaded clustering (SSCC) algorithm is proposed to identify the correct set of patterns/classes in a mislabeled dataset. The proposed algorithm employs the k-means and the k-medoids clustering algorithms and the Euclidean distance metric for classification. It eliminates the possible noisy-label information and the outliers using a novel cluster evaluation matrix (CEM) along with the cluster completeness score. The SSCC generates a cascaded classifier based on the initial dataset provided. The performance of the cascaded classifier is shown to be accurate and consistent despite variations in mislabeling of data and therefore, outperforming the typical supervised classifier such as SVM. The proposed method is developed as a part of a digital twin solution for a thermal power plant, for real-time detection and classification of coal type. The coal classification assists in real-time optimum operation of the plant and reduction in greenhouse emissions.  The algorithm was tested and validated against several public datasets including an industrial dataset.

\bibliographystyle{unsrt}  
\bibliography{mybibliography}  
\pagebreak
\appendix
\section{Appendix}

\begin{table}[h]
	\small
	\centering
	\caption{Threshold values for each dataset used.}
	\begin{tabular}{l c c c c c c c}
		\hline
		Datasets & Mislabeling (\%) & \multicolumn{3}{c} {Threshold Parameters} \\
		\cline{3-8}
		& & \multicolumn{2}{c}{$\lambda_{CEM}$} & \multicolumn{2}{c}{$\lambda_{CS}$} & \multicolumn{2}{c}{$\lambda_{OL}$}\\
		\cline{3-8}
		& & kmeans & kmedoids & kmeans & kmedoids & kmeans & kmedoids \\ 
		\hline
		\multirow{4}{*}{Coal Data} & 0 & 0.9 & 0.9 & 0.9 & 0.9 & 0.01 & 0.01\\
		& 10 & 0.9 & 0.9 & 0.9 & 0.9 & 0.01 & 0.01\\
		& 20 & 0.9 & 0.9 & 0.9 & 0.9 & 0.01 & 0.01\\
		& 30 & 0.9 & 0.9 & 0.9 & 0.9 & 0.01 & 0.01\\
		\hline
		\multirow{4}{*}{Ecoli Data} & 0 & 0.75 & 0.75 & 0.75 & 0.75 & 0.04 & 0.04\\
		& 10 & 0.75 & 0.75 & 0.75 & 0.75 & 0.04 & 0.04\\
		& 20 & 0.75 & 0.75 & 0.75 & 0.75 & 0.04 & 0.04\\
		& 30 & 0.75 & 0.75 & 0.75 & 0.75 & 0.04 & 0.04\\
		\hline
		\multirow{4}{*}{Wine Data} & 0 & 0.7 & 0.7 & 0.65 & 0.65 & 0.03 & 0.03\\
		& 10 & 0.7 & 0.7 & 0.65 & 0.65 & 0.03 & 0.03\\
		& 20 & 0.7 & 0.7 & 0.65 & 0.65 & 0.03 & 0.03\\
		& 30 & 0.7 & 0.7 & 0.65 & 0.65 & 0.03 & 0.03\\
		\hline
		\multirow{4}{*}{Eucalyptus Data} & 0 & 0.85 & 0.9 & 0.8 & 0.8 & 0.02 & 0.02\\
		& 10 & 0.85 & 0.9 & 0.8 & 0.8 & 0.02 & 0.02\\
		& 20 & 0.85 & 0.9 & 0.75 & 0.75 & 0.02 & 0.02\\
		& 30 & 0.85 & 0.9 & 0.75 & 0.75 & 0.02 & 0.02\\
		\hline
	\end{tabular}
	\label{table:threshold}
	
\end{table}

%
%
%
%

\end{document}